\documentclass[tinyml]{acmart}

\usepackage[ruled, vlined, noend]{algorithm2e}
\usepackage{amsmath}
\usepackage[nameinlink]{cleveref}
\usepackage{comment}
\usepackage{enumitem}
\usepackage{graphicx}
\usepackage{subfig}
\usepackage{multicol}
\usepackage{float}
\usepackage[export]{adjustbox}
\AtBeginDocument{%
  \providecommand\BibTeX{{%
    \normalfont B\kern-0.5em{\scshape i\kern-0.25em b}\kern-0.8em\TeX}}}

\setcopyright{rightsretained}
\copyrightyear{2023}
\acmYear{2023}

\setcopyright{none}
\begin{document}

\title{MEMA Runtime Framework: Minimizing External Memory Accesses for TinyML on Microcontrollers}

\author{Andrew Sabot}
\authornote{Both authors contributed equally to this research.}
\email{asabot@g.harvard.edu}
\affiliation{%
  \institution{Harvard Universtiy}
  \streetaddress{P.O. Box 1212}
  \city{Cambridge}
  \state{MA}
  \country{USA}
  \postcode{43017-6221}
}
\author{Vikas Natesh}
\authornotemark[1]
\email{vnatesh@g.harvard.edu}
\affiliation{%
  \institution{Harvard Universtiy}
  \streetaddress{P.O. Box 1212}
  \city{Cambridge}
  \state{MA}
  \country{USA}
  \postcode{43017-6221}
}

\author{H.T. Kung}
\email{kung@harvard.edu}
\affiliation{%
  \institution{Harvard Universtiy}
  \streetaddress{P.O. Box 1212}
  \city{Cambridge}
  \state{MA}
  \country{USA}
  \postcode{43017-6221}
}

\author{Wei-Te Ting}
\email{weiteting@g.harvard.edu}
\affiliation{%
  \institution{Harvard Universtiy}
  \streetaddress{P.O. Box 1212}
  \city{Cambridge}
  \state{MA}
  \country{USA}
  \postcode{43017-6221}
}
\renewcommand{\shortauthors}{Names Redacted}
\begin{abstract}
We present the MEMA framework for the easy and quick derivation of efficient inference runtimes that \underline{m}inimize \underline{e}xternal \underline{m}emory \underline{a}ccesses for matrix multiplication on TinyML systems.
The framework accounts for hardware resource constraints and problem sizes in analytically determining optimized schedules and kernels that minimize memory accesses.
MEMA provides a solution to a well-known problem in the current practice, that is, optimal schedules tend to be found only through a time consuming and heuristic search of a large scheduling space.
We compare the performance of runtimes derived from MEMA to existing state-of-the-art libraries on ARM-based TinyML systems.
For example, for neural network benchmarks on the ARM Cortex-M4, we achieve up to a 1.8x speedup and 44\% energy reduction over CMSIS-NN.
\end{abstract}

\keywords{tinyML, matrix multiplication, arithmetic intensity, outer product, memory access, computation scheduling, neural networks}

\maketitle
\pagestyle{plain}

\section{Introduction}
\label{sec:introduction}

Small Internet of things (IoT) devices have become increasingly common, and used in a growing number of fields including healthcare and consumer products\cite{mitre_tinyml}.
As these devices grow more popular, the amount of data collected increases, driving the demand for computation and machine learning (ML), or TinyML \cite{lin2020mcunet}, on these systems.
However, due to size and power constraints, these devices are heavily limited in their available memory, bandwidth, and computation power.
In addition, TinyML devices may be deployed as a distributed system (e.g., for AR/VR workloads \cite{gomez2022distributed}) with limited communication bandwidth.
Thus, there is a need to support efficient computation for machine learning applications on a wide variety of system configurations for such devices.

Machine learning models involve operations such as convolutional layers, fully connected layers, ReLU, max pooling, and batchnorm.
Computations in convolutional and fully connected layers of convolutional neural networks (CNNs) and attention layers of transformers may be implemented as matrix multiplications (MMs) (see, e.g., \cite{warden2015why, mcdanel2019full}).
As a result, MM makes up a substantial amount of the runtime.
When MMs do not fit entirely in the local memories of computing hardware (e.g., registers and caches on CPUs), additional data transfers to and from external memory are needed.
This additional IO increases energy consumption and latency.

TinyML hardware characteristics and capabilities can vary significantly, so optimal choices of kernels and schedules may differ for the same problem across multiple devices.
Consequently, developing a runtime framework that can find optimal schedules for the wide variety of TinyML systems is key for efficient inference.


To characterize the capabilities of system architectures, roofline plots, (see, e.g., \cite{roofline}) are commonly used.
The slope of the slanted line is the memory IO bandwidth and the horizontal line is the peak computation throughput.
The horizontal position of the ridge point represents the minimum \textit{arithmetic intensity} (arithmetic operations performed per IO operation) required to achieve the peak computation throughput offered by the hardware.
The optimal schedule for minimizing memory accesses is a function of the roofline plot. 
For example, some microcontroller units (MCUs), e.g., the ARM Cortex-M4, have relatively high memory bandwidths compared to their computation power \cite{gables_roofline}.
For these systems, the  ridge point is associated with a smaller arithmetic intensity.



\begin{figure}[h]
    \centering
        \quad
        \includegraphics[width=.75\linewidth]{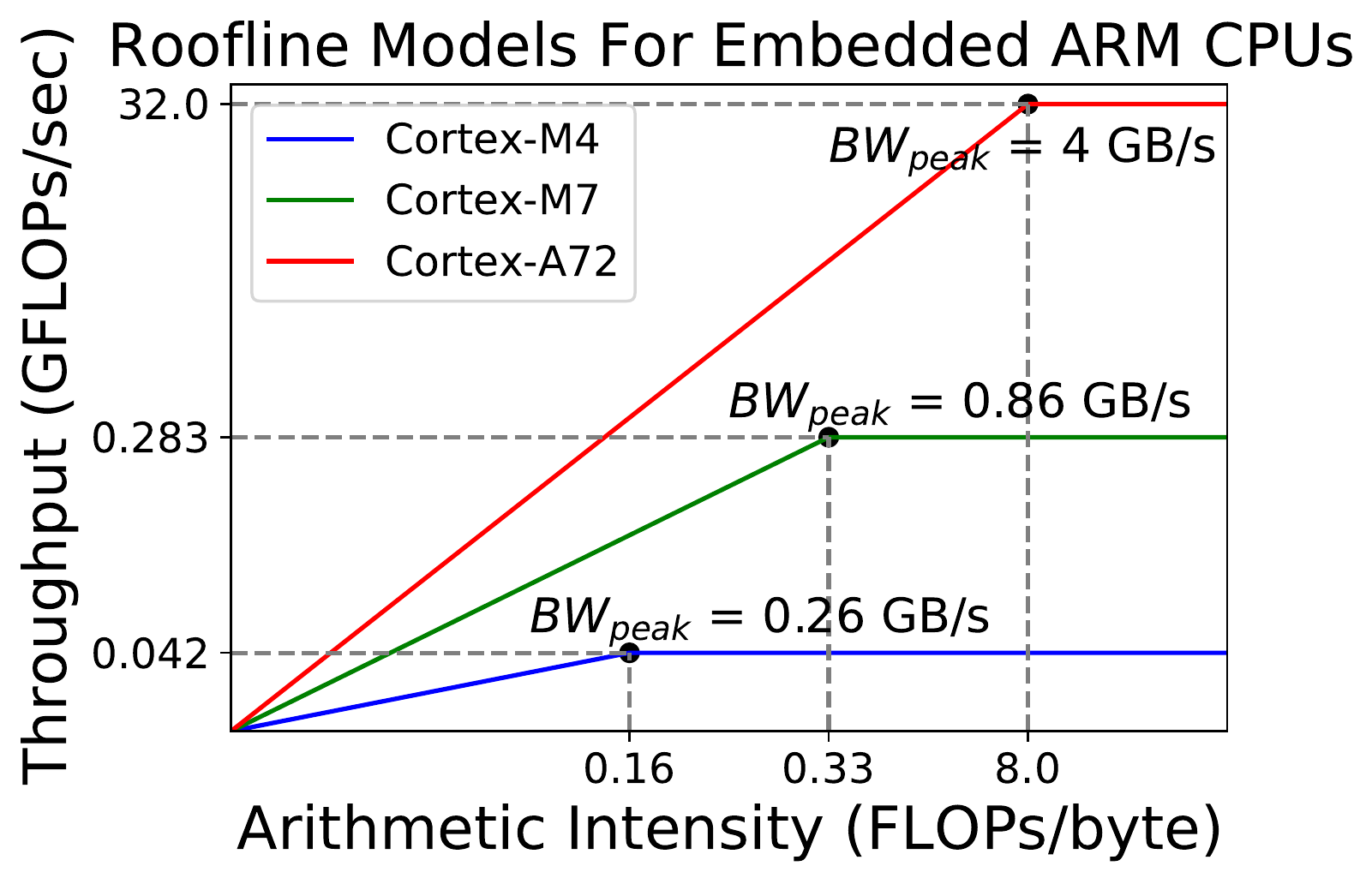}

    \caption{
    Roofline plots for ARM Cortex-M4, M7, and A72.
    The ridge point of each device is denoted as a black dot.
    Systems with a ridge point further to the right are bandwidth-bound, indicating that a higher arithmetic intensity will increase throughput.
    Conversely, systems with a ridge point further to the left have large memory bandwidths relative to computation throughput.
    }
    \label{fig:intro_preview_roofline}
\end{figure}

In this paper, we present the MEMA runtime framework for producing efficient inference runtimes for TinyML that minimize external memory accesses.
The framework analyzes the hardware and computational problem to derive IO-efficient runtime schedules.
We use the roofline model to reason about the MEMA approach, focusing on tiny devices with small local memories.
For the three architectures evaluated in this paper (ARM Cortex-M4, M7, and A72), we present their roofline plots in \Cref{fig:intro_preview_roofline}.
The MEMA framework schedules computations to increase arithmetic intensity for two reasons: 1) increasing computation throughput when left of the ridge point and 2) decreasing the required IO bandwidth, and thereby energy for memory IO when right of the ridge point.

\begin{figure}[h]
    \centering
    \includegraphics[width=1.1\linewidth]{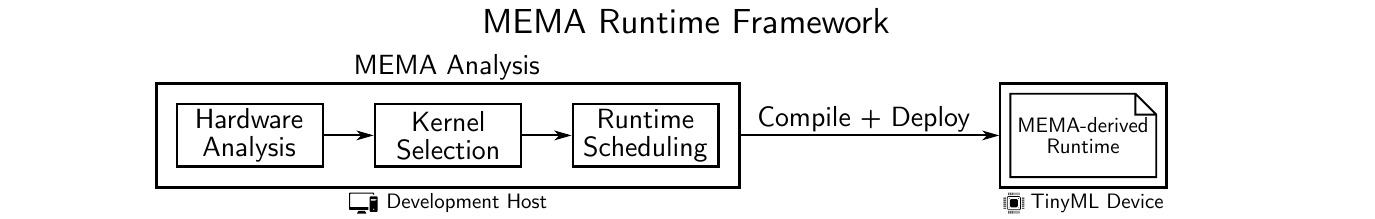}
    \caption{Overview of MEMA runtime framework.}
    \label{fig:mema_overview}
\end{figure}

The approach of selecting the appropriate library and kernels based on hardware characteristics is a common practice on CPUs \cite{smith2019momms}. 
However, these techniques have not yet been applied to MCUs. 
We cannot directly transfer decisions in the CPU domain to the MCU domain since the analytical arguments are very different due to disparities in the hardware (e.g., differences in the memory hierarchy and available instructions). 
As a result, in this paper, we introduce and validate the backbone analysis for MCUs.


\Cref{sec:background_and_related_work} provides background on matrix multiplication and introduces related works.
\Cref{sec:mema_framework} overviews the main objectives of the MEMA runtime framework and describes how streaming as a scheduling technique maximizes data reuse in local memory so derived runtimes can operate on TinyML devices with severely constrained memory footprint.
\Cref{sec:mema_analysis} breaks down the general MEMA analysis process for a tiled MM.
\Cref{sec:mema_runtime} gives an example of how to use the analysis from \Cref{sec:mema_analysis} to derive a MEMA schedule on real hardware.
\Cref{sec:experiment_methodology} describes the methodology and benchmarks used in our evaluation of MEMA.
\Cref{sec:results} demonstrates the performance of MEMA runtimes compared to the state-of-the-art libraries on several TinyML devices.

This paper makes the following contributions:
\begin{itemize}[noitemsep]
    \item MEMA runtime framework that reduces total memory IO requirements and hides IO times in computation times for generated runtimes.
    To our knowledge, MEMA is the first such framework aiming to ease the challenge of designing such schedules for TinyML systems. 
    \item MEMA formalizes the streaming framework to support the proper shaping and allocation of input and result tiles ($A$, $B$, and $C$) in local memory on MCUs with highly constrained memory systems (\Cref{sec:mema_analysis})
    \item MEMA analytically derives optimal tiling for multiple bit widths without the need for searching, as shown in \Cref{fig:mematinyml}.

    \item Empirical results demonstrating MEMA runtime performance on real hardware (\Cref{sec:results}).
\end{itemize}

\section{Background and Related Works}
\label{sec:background_and_related_work}

\subsection{Partitioning a Matrix Multiplication into Computation Blocks}
\label{subsec:mm_partitioning}
\begin{figure}
    \centering
    \includegraphics[width=\linewidth]{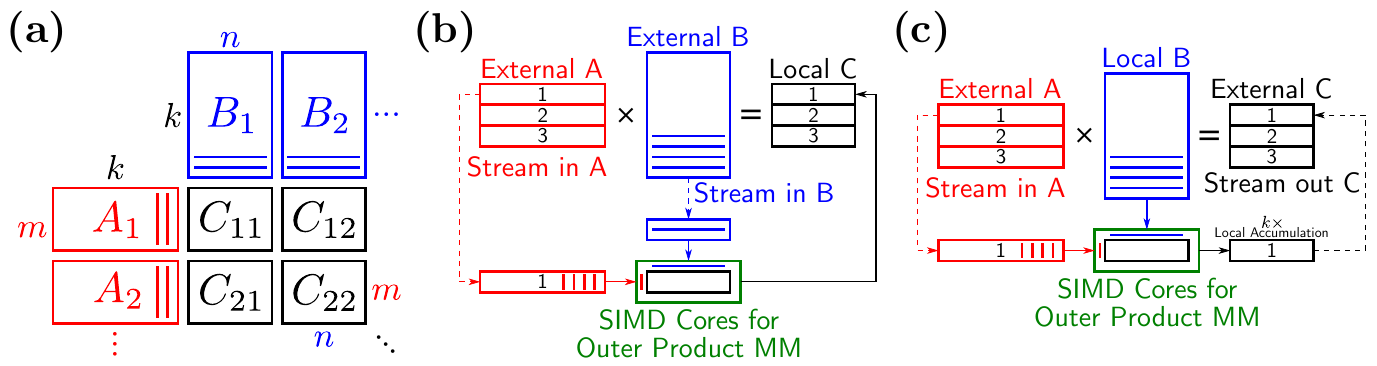}

    \caption{(a) MM for each value of $t_3$ in \Cref{algo:m-first} between $M \times k$ and $k \times N$ matrices using $m \times k \times n$ computation blocks.   
    $A_i$ and $B_i$ are tiles of $A$ and $B$, with lines indicating column and row vectors, respectively. 
    $C_{ij}$ are tiles of $C$.
    (b) We can hold rows of $C$ stationary in the local memory and stream in rows of $A$ and $B$.
    (c) By streaming computed rows of $C$ to external memory, we can hide the IO time within the compute time.
    When $k$ is sufficiently large, the computation time for each row of $C$ exceeds the time to write the row back to external memory.
    In contrast to (b), this scheme need only hold one row of $C$ as intermediary results.
    (b) and (c) are both examples of schedules that can be generated by MEMA.
    }
    \label{fig:tile_mm_and_streaming}
\end{figure}

For MMs where $A$, $B$, and $C$ do not all fit in local memory, we must partition, or tile, the MM into smaller $m \times k \times n$ computation blocks so the inputs to each block fit in local memory (\Cref{fig:tile_mm_and_streaming}a).

\begin{algorithm}
\SetInd{0.25em}{0.5em}
// loops on $M/m \times K/k \times N/n$ computation blocks\\
\For{$t_3 = 0 \to K, \quad t_3 += k$}{
    \For{$t_2 = 0 \to N, \quad t_2 += n$}{
        \For{$t_1 = 0 \to M, \quad t_1 += m$}{
            // loops for a single computation block\\
            \For{$k^*= t_3 \to t_3+k, \quad k^*++$}{
                \For{$j = t_2 \to t_2+n, \quad j++$}{
                    \For{$i = t_1 \to t_1+m, \quad i++$}{
                    $C[i+t_1\cdot m][j+t_2\cdot n] \longleftarrow C[i+t_1\cdot m][j+t_2\cdot n]  + A[i+t_1\cdot m][k^* +t_3\cdot k] \cdot B[k^* +t_3\cdot k][j+t_2\cdot n]$\;
                    
                    }
                }
            }
        }
    }
}
\caption{$M$-first MM using $m\times k \times n$ blocks.}
\label{algo:m-first}
\end{algorithm}

\Cref{algo:m-first} describes a block-partitioned MM where, for each value of $t_3$ in the outermost loop, an outer-product-based MM between two submatrices ($M \times k$ and $k \times N$ matrices) is performed. 
We denote this scheme as $M \rightarrow N \rightarrow K$ to reflect the order of the nested loops.
For this outer product MM, the scheme performs $M$-first block computations ($t_1$ in the innermost loop) as opposed to $N$-first block computations. 
There are \textbf{six schemes} for scheduling computation blocks:
$M \rightarrow N \rightarrow K$, $N\rightarrow M \rightarrow K$, $M \rightarrow K \rightarrow N$, $N \rightarrow K \rightarrow M$, $K \rightarrow M \rightarrow N$, and $K \rightarrow N \rightarrow M$.

\subsection{Stationary and Streamed Data}
\label{subsec:operand_streaming}

For each MM computation, $A$, $B$, and $C$ tiles must be loaded from external memory.
When data is kept stationary in local memory, we are able to reduce the number of tiles fetched from external memory.
We designate data kept in local memory for multiple computations as stationary (e.g., $B$ matrix in \Cref{fig:tile_mm_and_streaming}b).
Data fetched from external memory and used once before being evicted (e.g., $A$ matrix in \Cref{fig:tile_mm_and_streaming}b) is considered ``streamed``.
Accumulated results may be streamed out to external memory (e.g., computed rows of $C$ in \Cref{fig:tile_mm_and_streaming}b).
Streaming techniques have been used in prior works such as \cite{goto2008anatomy}, but our novelty lies in using streaming when automating runtime schedule design.

Using MEMA we can decrease the total amount of data streamed to/from external memory.
When the computation time is larger than the streaming time for streamed data items, we are able to hide the streaming time (\Cref{fig:tile_mm_and_streaming}) and decrease the required local memory size.
The IO to write back a portion of $C$ can be overlapped with computation of the next portion of $C$. 
We assume that there is sufficient bandwidth to also stream in matrix $A$ from external memory (as is the case of the Cortex-M4, with a ridge point further to the left in \Cref{fig:intro_preview_roofline}c).
The streaming method works similarly when $A$, $B$, or $C$ is kept stationary.
For all three methods, smaller tiles of $C$ are accumulated in-place to reduce data movement.
If external bandwidth is insufficient, compression techniques, such as \cite{8675200} can be applied to the data being streamed in.

\subsection{Related Works}
CMSIS-NN \cite{lai2018cmsisnn} is a set of kernels for common neural network operations, focusing on performance in throughput and latency, and minimizing memory footprints of neural networks on ARM Cortex-M processors.
By using inner products and fixing the loop order, CMSIS-NN is not able to maximize data reuse. 


In addition to CMSIS-NN, recent frameworks such as MCUNet \cite{lin2020mcunet} perform loop tiling and unrolling for neural network layer operations such as convolution.
MCUNet only considers the problem size and available local memory when choosing tile sizes and does not explicitly minimize memory bandwidth usage.
In contrast, our work uses both local memory and roofline characteristics of the specific device to derive efficient tile sizes.
In addition, we leverage outer products and loop reordering to increase data reuse, reducing latency and energy consumption (\Cref{fig:mematinyml}).

ARM provides two libraries for MM on ARM Cortex-A devices: ARM Compute Library for machine learning (ARMCL \cite{armcl}) and ARM Performance Library (ARMPL \cite{armpl}).
While ARMPL and ARMCL use outer product-based methods for MM, their computation schedule and tiling strategy is based on Goto's algorithm \cite{gotomain}, which does not minimize IO bandwidth usage.
By minimizing external accesses, we are able to outperform ARMCL and ARMPL (see \Cref{fig:mema_a72}).
GOTO's algorithm \cite{gotomain} is a classical algorithm based on data streaming for high-performance MM on CPUs, underlying OpenBLAS \cite{xianyi2012openblas} and Intel MKL \cite{intelmkl}.
CUTLASS \cite{nvidia_developer_blog_2020} is an open-source C\texttt{++} CUDA BLAS library for Nvidia GPUs.
CUTLASS is similar to our work in using outer product formulations for MM.

TensorFlow Lite Micro (TFLM) \cite{david2021tensorflow} is an ML inference framework for deep learning on embedded systems.
Hardware vendors can contribute their own kernels to TFLM, allowing programmers to deploy ML models to many architectures.
Currently, TFLM uses the CMSIS-NN library when benchmarking embedded hardware platforms \cite{banbury2021mlperf}.
In this paper, we demonstrate our kernel improves energy usage, bandwidth usage, and computation throughput over the CMSIS-NN kernel for TFLM.


\section{MEMA Framework}
\label{sec:mema_framework}
The MEMA runtime framework combines hardware characteristics, problem size information (e.g., input matrix sizes for MM), IO analysis, and scheduling to produce an efficient inference runtime.
In this section, we introduce the MEMA framework objectives.
We also describe how MEMA leverages streaming to maximize data reuse and overlap compute and IO times when partitioning MM computations with inputs that do not fit in local memory.

TinyML devices are constrained in memory sizes and speeds due to form factor and limited power budgets.
The MEMA framework addresses these challenges through its objectives: reducing the number of external memory accesses and hiding IO time within compute time via streaming that efficiently utilizes available local memory.
By reducing memory accesses, the framework may reduce energy consumption of machine learning inference tasks on TinyML devices.
This allows MEMA to overlap IO with compute and derive runtimes that are not bottlenecked by IO.

\subsection{Overview of the MEMA Framework}
Given an MM $C = C + A \times B$, where $A$ is $M \times K$ and $B$ is $K \times N$, the MEMA framework can automatically derive optimal schedules in minimizing external memory accesses, subject to local memory size, using techniques that maximize data reuse: (1) tile shaping, (2) matrix operand streaming and (3) loop order selection.

The framework first uses information about the target hardware, such as register count, local memory size, and external memory bandwidth to determine the optimal tile sizes (\Cref{subsec:tile_selection}).
Tile sizes are automatically derived to maximize the arithmetic intensity of each tile multiplication (with the goal of increasing computation throughput and reducing energy consumption for IO).
Based on the tile sizes, MEMA can select kernels tailored to the target platform from existing libraries or generate optimized outer product kernels.
Then, by accounting for potential tile sizes and MM problem size, a schedule that minimizes external IO (\Cref{sec:loop_tile_IO}) and decreases the required IO bandwidth is selected.
The kernels and schedules are then compiled into a MEMA runtime and deployed to the TinyML device, as depicted in \Cref{fig:mema_overview}.

\section{MEMA Analysis for Runtime Derivation}
\label{sec:mema_analysis}
In this section we cover the MEMA analysis for MM on two hardware platforms: a single core MCU and a multi-core IoT device.
Our analysis is limited to MM operations, but the framework may be extended to other multi-loop reduction operations with static loop bounds such as direct convolution and tensor contractions.
The MEMA analysis starts by analytically deriving a tile size that maximizes arithmetic intensity for a given hardware.
Then using the derived tiling, we select a loop order which maximizes data reuse between successive tile computations.
We show that the external memory IO associated with streaming is determined by both tile size and MM problem size.
Meanwhile, the MM problem size alone determines the IO for the  stationary data.
MEMA chooses the loop order with the lowest combined streaming and stationary IO.

\subsection{Deriving Tile Sizes for MCUs}
\label{subsec:tile_selection}
Consider an outer product shown in \Cref{fig:tile_mm_and_streaming}a between a $m\times1$ vector of $A$ and $1\times n$ vector of $B$ to produce a stationary $m\times n$ tile of $C$.
For simplicity of discussion, we only count multiplications, excluding additions.
To maximize reuse, we maximize the ratio of computation ($mn$ multiplications) to IO ($m+n$ input values) i.e., the arithmetic intensity $\frac{mn}{m+n}$.
This ratio is maximized when we choose a square tile $m=n=t$.
Assume we have an MCU with 36 registers of local memory available for data reuse.
All outer product inputs ($m + n + mn$ values) must fit in local memory, i.e., $2t + t^2 \leq 36$.
Solving for $t$ shows the optimal tile size of $C$ for this MCU is $5\times5$. 

The selected $C$ tile in this example will act as an intermediate local store to support the input streaming with computation when the scheme of \Cref{fig:tile_mm_and_streaming}c is used.
While we derive tile dimensions via a simple arithmetic intensity argument here, other tiling techniques may be used \cite{kung2021cake,bondhugula2008automatic} with MEMA. 


\subsection{Deriving Tile Sizes for Multicore IoT Devices}
\label{subsec:cake}

Unlike MCUs, IoT devices such as \cite{arm72,rossi2015pulp,GAP8} contain multiple low-power RISCV or ARM cores as well as multiple levels of memory. 
Since we have to tile at multiple memory levels, the number of possible schedules is much larger. 
For example, on a device with 3 memory levels, we have the choice of 6 MM loop orders at each level for a total of $6^3 = 216$ possible loop orderings. 
In addition, tiles at each memory level must be properly sized according to the available local memory, bandwidth, and number of cores.

Instead of searching a large space of schedules for the optimal tiling as in \cite{dory, acmtc}, we use CAKE \cite{cake}, a multi-core matrix multiplication tiling and scheduling algorithm that utilizes \textbf{constant-bandwidth (CB) blocks} in computation partitioning and block scheduling. 
A CB block is a block of computation that, when computed from within a local memory, the required off-chip bandwidth is constant, even when utilizing additional cores. 
CAKE controls the arithmetic intensity of CB blocks by adjusting the CB block shape (i.e., aspect ratios) and size according to available off-chip DRAM memory bandwidth, number of cores, and available local memory.
Using CAKE tiling we can increase the use of available computing power without requiring a comparable increase in off-chip memory bandwidth.
CAKE does require more local memory when increasing the number of cores, but local memory size is often sufficient in comparison to off-chip bandwidth.

Suppose we want to grow the number of cores by a factor $p$. CAKE reshapes the computation block from $m\times k \times n$ (shown in \Cref{fig:tile_mm_and_streaming}) to the shape $pm \times k \times pn$. 
CAKE also holds the large $C$ tile, with dimensions $pm \times pn$, stationary in on-chip memory while streaming in the smaller $pm \times k$ and $k \times pn$ tiles of $A$ and $B$, respectively. 
Let $p\cdot f$ be the peak FLOPs of the system where $f$ is the single core peak. 
To compute the CB block in local memory, CAKE performs $pm\cdot k \cdot pn$ MAC operations in time $T=\frac{pm\cdot k \cdot pn}{p\cdot f}$ using $IO = pmk + pnk$ off-chip memory accesses.
The required off-chip bandwidth is then:
{\small
\[
BW_{off-chip} = \frac{IO}{T} = \frac{p\cdot k\cdot (m+n)}{pm\cdot k \cdot pn} \cdot\! p \cdot\! f = \frac{m+n}{m n}\cdot\! f
\] }

Note that CAKE's bandwidth usage is constant regardless of the number of cores because the $p$ factors cancel out. Given CAKE's CB block shape, we can directly solve for the optimal tile sizes by choosing $m$, $k$, and $n$ such that $A$, $B$, and $C$ tiles fit in local memory ($LM$), i.e., $pmk + kpn + p^2mn \leq LM$. Here, $LM$ is on-chip memory shared by all the cores, each of which computes a single $m\times k \times n$ sub-block at a time. Without loss of generality, we may continue to tile the $m\times k \times n$ sub-block according to the available registers or local memory private to each core (see \Cref{sec:results}). 

\subsection{Loop Order for Inter-Block Computations}
\label{subsec:mema_loop_order}
Computing computation blocks with different loop orders may result in different total external memory accesses for the same MM.
CAKE does not reorder the loops to minimize external memory accesses, instead it only uses a $K$-first scheduling of blocks ($K$-dimension as the inner loop, keeping partial results stationary). 
Total IO varies since loop order determines the streaming pattern.
For skewed matrix shapes, a partial result stationary schedule may not minimize off-chip memory accesses.
MEMA improves upon this by selecting the loop order that minimizes total external memory accesses, even if the order does not keep partial results stationary.


For a partitioned MM between an $M \times K$ matrix $A$ and a $K \times N$ matrix $B$ (described in \Cref{subsec:mm_partitioning}) the total IO for a given loop order can be computed as the sum of streaming and stationary IO:
\begin{equation*}
    (\# \text{ of blocks}) \cdot (IO_{streaming} \text{ per block}) + 
    IO_{stationary}
\end{equation*}
\noindent
For an $N$-first schedule (stationary $A$ tiles), our total IO is:
{\small
\begin{align*}
    IO_{N\text{-first}} &= \frac{MKN}{mkn}(2mn + nk) + \frac{M}{m} \cdot \frac{K}{k}(mk) = MKN \left(\frac{1}{m} + \frac{2}{k}\right) + MK
\end{align*}}
For an $M$-first schedule (stationary $B$ tiles), our total IO is:
{\small
\begin{align*}
    IO_{M\text{-first}} &= \frac{MKN}{mkn}(mk + 2mn) + \frac{K}{k} \cdot \frac{N}{n}(kn) = MKN \left(\frac{2}{k} + \frac{1}{n} \right) + KN
\end{align*}}
For an $K$-first schedule (stationary $C$ tiles), our total IO is:
{\small
\begin{align*}
    IO_{K\text{-first}} &= \frac{MKN}{mkn} (mk + kn) + \frac{M}{m} \cdot \frac{N}{n}(2mn) = MKN \left(\frac{1}{m} + \frac{1}{n} \right) + 2MN
\end{align*}}
\noindent
The 2 factor from $2mn$ reflects reading \textit{and} writing partial $C$ tiles.

\subsection{Choosing Loop Ordering Based on Tile Dimensions and Input Matrix Sizes}
\label{sec:loop_tile_IO}
Using the IO equations in \Cref{subsec:mema_loop_order}, MEMA selects a schedule which minimizes IO.
Tile size determines the number of memory accesses for streaming.
The size of $A$ and $B$ determines the memory accesses for stationary data.
A loop order that minimizes external memory accesses may be selected by calculating the memory accesses for each of the 6 possible loop orders given in \Cref{subsec:mm_partitioning}.
The rest of this section gives two examples of MM loop order selection.

Consider an MM where $M=K=N$ computed with non-square tiles ($n > k = m$).
The $B$ and $C$ tiles are larger and thus would require more IO to stream. 
In an $N$-first loop order, an $A$ tile is kept stationary while $B$ and $C$ tiles are streamed.
However, we can reduce IO by using an $M$-first or $K$-first order and streaming $A$ tiles. 
In this example, the $M$-first and $K$-first orders are equivalent as all other dimensions are equal.

Now consider an MM between $A$ and $B$ with $M > N$ and $M > K$.
Suppose the MM is computed using square tiles ($n = m = k$).
Computing the MM using the $N$-first requires more total external memory accesses than the $M$-first schedule.
In comparison, the $M$-first order has less IO since $B$ tiles can be reused $\frac{M}{m}$ times, which is more than other directions because $M > N$ and $M > K$.
The streaming term of the IO function is independent of loop order when tile sizes are equal.


\section{MEMA Derivation for Real Hardware}
\label{sec:mema_runtime}

\subsection{Evaluation Hardware Characteristics}
\label{subsec:evaluation_hardware}
In this and subsequent sections (\Cref{sec:experiment_methodology} and \Cref{sec:results}) we evaluate MEMA's performance against state-of-the-art libraries for MM on multiple MCUs representative of popular embedded platforms: Arduino Nano 33 BLE (ARMv7E-M Cortex-M4), STM32F767ZI Nucleo (ARMv7E-M Cortex-M7), and Raspberry Pi 4 Model B (ARMv8-A Cortex-A72).
MEMA is implemented using C++ and compiler intrinsics on each respective platform (open-sourced at \url{https://github.com/vnatesh/MEMA-MM}).
The Cortex-M4 and M7 have very limited local memory: only 32 floating-point registers and 16 registers for DSP-accelerated fixed-point arithmetic.
The Cortex-M4 has a 256KB external memory (SRAM) and 1MB of flash memory for storing additional data and code.
The Cortex-M7 has roughly double the SRAM and flash of the M4. 
Meanwhile, the Cortex-A72 has more compute and memory resources than the Cortex-M4 and M7 but is still limited by DRAM bandwidth, local memory size, and local memory bandwidth relative to computation power (see \Cref{fig:intro_preview_roofline}).

\subsection{MEMA Schedule Derived on Cortex-M4}
\label{subsec:hardware_schedule_derivation}
In this section, we apply MEMA's analysis to an ARM Cortex-M4 in minimizing memory accesses for an MM problem, given $M$, $K$, $N$.
After determining an optimal tile size based on the device's local memory size and hierarchy (\Cref{subsec:tile_selection}), we derive conditions for when to use any of the six computation block schedules (\Cref{subsec:mema_loop_order}).
Consider, e.g., the choice between $M$ and $K$-first schedules. 
To minimize IO, we should use the $M$-first schedule only when $IO_{M\text{-first}} \leq IO_{K\text{-first}}$ and $IO_{M\text{-first}} \leq IO_{N\text{-first}}$.
Using the equations in \Cref{subsec:mema_loop_order}, we obtain the following inequalities as a condition for choosing the $M$-first schedule:
{\small
\begin{align*}
  MKN \left(\frac{2}{k} + \frac{1}{n} \right) + KN &\leq MKN \left(\frac{1}{m} + \frac{1}{n} \right) + 2MN  \\
  MKN \left(\frac{2}{k} + \frac{1}{n} \right) + KN &\leq MKN \left(\frac{2}{k} + \frac{1}{m} \right) + MK \\
\implies K \leq \left(\frac{2M}{1 + M(\frac{2}{k} - \frac{1}{m})} \right) \hspace{0.5em} &, \hspace{0.5em}  N \leq \left(\frac{M}{1 + M(\frac{1}{n} - \frac{1}{m})}\right)
\end{align*} }

\begin{figure}
    \centering
    \includegraphics[width=0.45\linewidth]{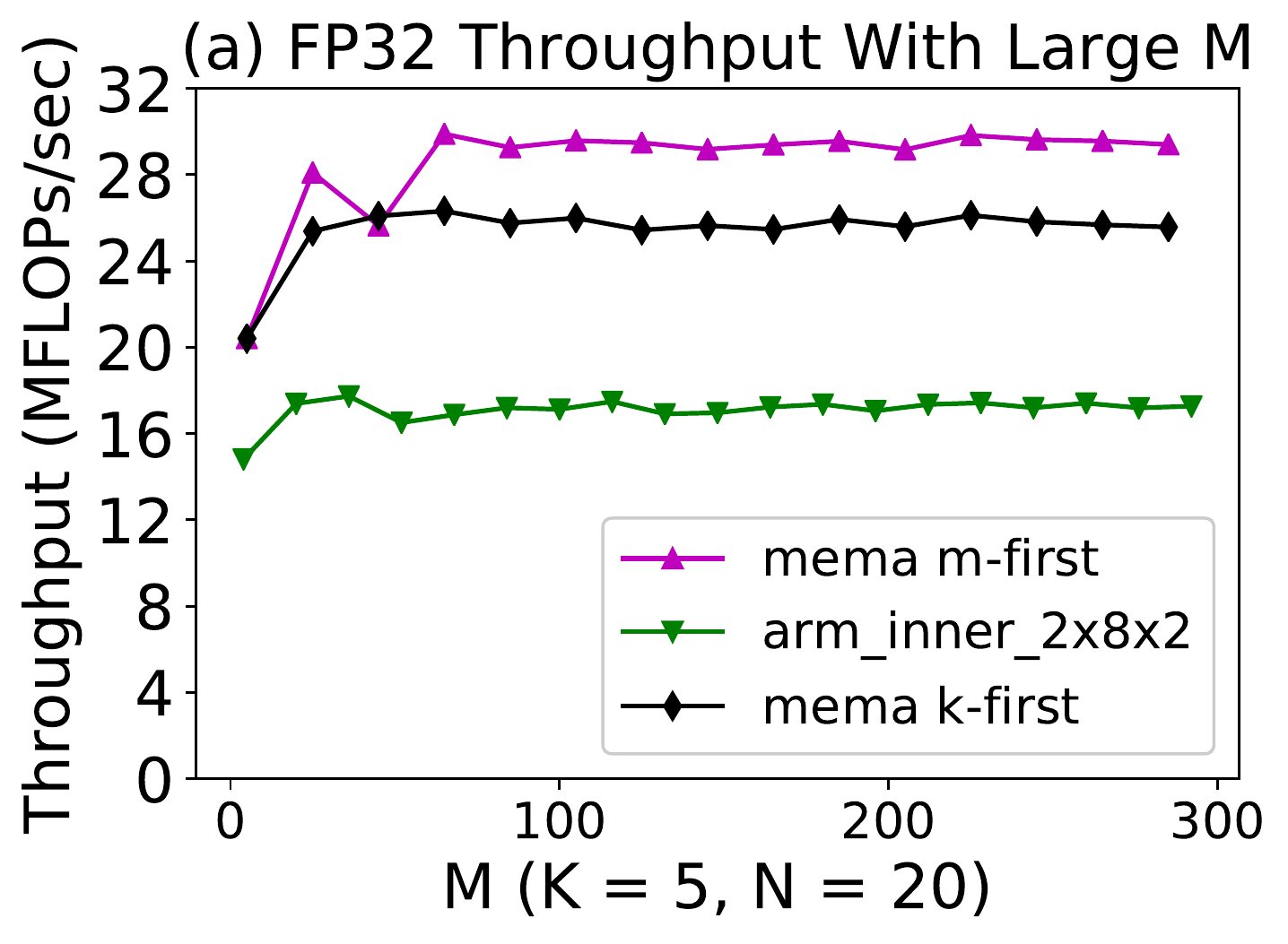}
    \qquad
    \includegraphics[width=0.45\linewidth]{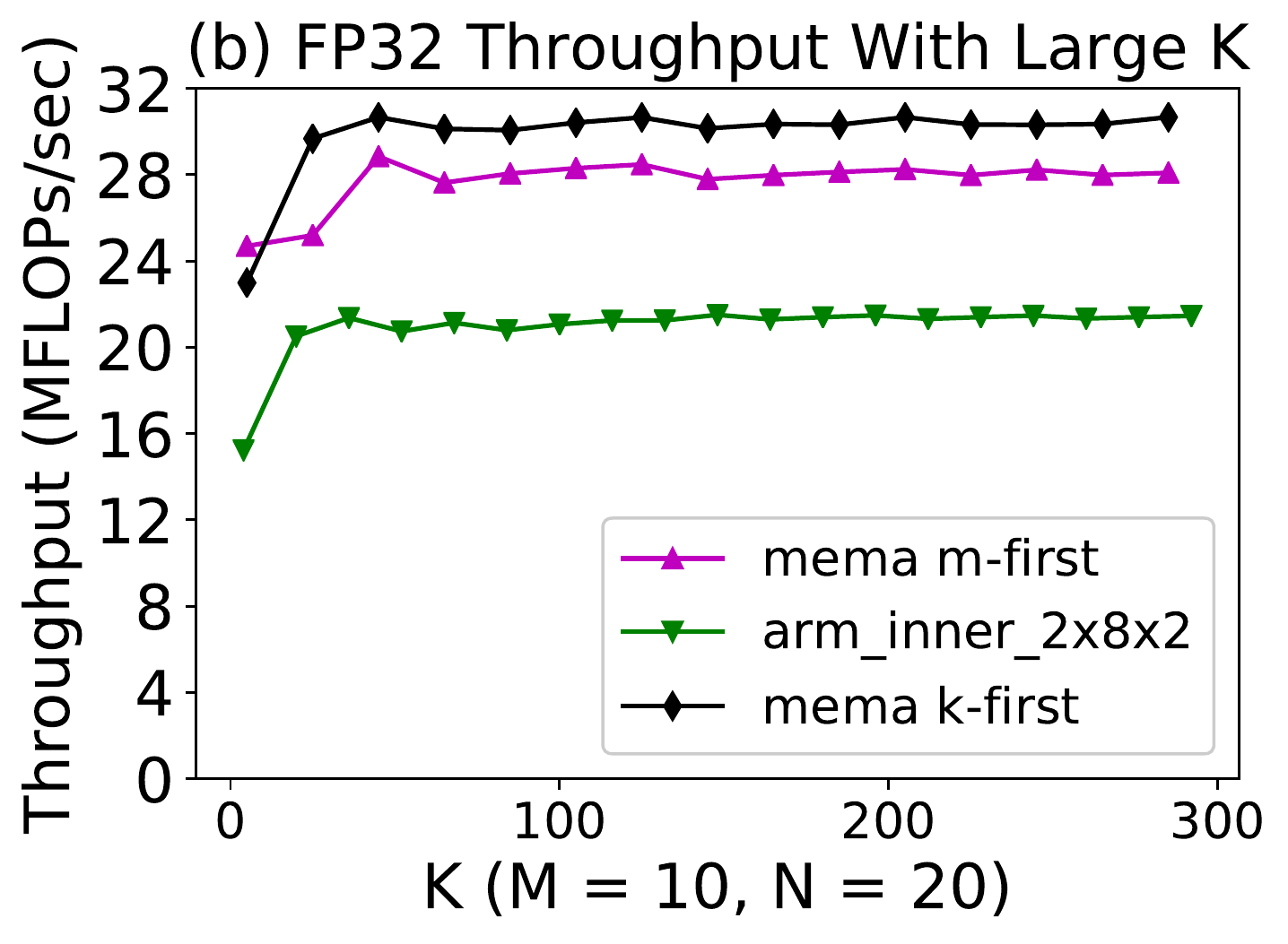}
    \caption{
    MEMA performance on Cortex-M4 for various MM problems. 
    By scheduling to minimize IO, MEMA maximizes throughput for matrices with skewed shapes.
    (a) $M$-first schedule outperforms $K$-first when $K=5$, as predicted by MEMA's analysis (\Cref{subsec:hardware_schedule_derivation}). 
    Similarly, (b) shows the $K$-first schedule outperforms $M$-first for $K > 10$.
    The MEMA schedule outperforms the ARM library (\texttt{arm\_inner\_2x8x2}).
    }
    \label{fig:memaskew}
\end{figure}

Since the Cortex-M4 has 36 registers available for local data reuse, suppose we use a square $5\times 5$ tile for stationary data as derived in \Cref{subsec:tile_selection} i.e., $m=k=n=5$. 
Then, the analysis above suggests we should use the $M$-first schedule when $K$ is roughly $\leq10$ and $M\geq N$, since in this case the above two inequalities hold. 
We confirm this choice empirically in \Cref{fig:memaskew}a which shows the $M$-first schedule outperforming $K$-first on the Cortex-M4 when $K=5$.
Similarly, \Cref{fig:memaskew}b shows that as $K$ increases beyond 10 with $M$ and $N$ fixed, the $K$-first schedule outperforms $M$-first. We may  decide between $N$ and $K$-first schedules via $IO_{N\text{-first}} \leq IO_{K\text{-first}}$, yielding inequalities analogous to those above. 

\section{Neural Network Benchmarks and Energy Measurement Methodology}
\label{sec:experiment_methodology}

\paragraph{Neural Network Benchmarks}
We measure computation throughput (FLOPs/sec), external memory IO (bytes), and energy usage (joules) during MMs between weight and data matrices for the layers of neural network models from various benchmarks. 
On the Cortex-M4 and M7, we use the MLPerf Tiny Benchmark \cite{banbury2021mlperf}, composed of models for tasks such as keyword spotting, visual wake words, and image recognition. 
Using matrices from the benchmark, we extract dimensions for weight and data matrices in each layer after applying \textit{im2col} \cite{im2col}. 
For the ARM Cortex-A72, we use transformer model matrices provided by the Deep Learning Matrix Collection (DMLC) \cite{dlmc}.
Packing overhead is expensive for the matrix sizes evaluated, so our kernels on the M4 and M7 are packing-free.

\begin{figure}
    \centering
    \includegraphics[width=0.9\linewidth]{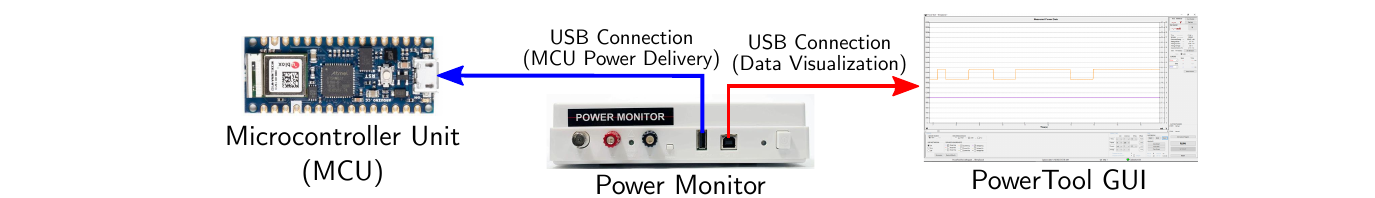}
    \caption{
    Power monitor setup for collecting data on a USB-connected MCU.
    The monitor reports real time voltage and current via the PowerTool GUI
    \cite{monsoonsolutionsinc.low}. 
    }
    \label{fig:power_monitor}
\end{figure}

\paragraph{Energy Measurement Setup}
To measure energy consumption, we follow the MLPerf Tiny benchmark guidelines \cite{banbury2021mlperf}.
We use a Monsoon Solutions Low-Voltage Power Monitor \cite{monsoonsolutionsinc.low}, which has a current resolution of 50 $\mu$A and voltage resolution of 125 $\mu$V.
The power monitor samples voltage and current every 200 $\mu$s and reports the data via the PowerTool GUI on a host desktop.
MCUs are connected to the power monitor USB channel with the USB passthrough feature enabled to allow communication with the Power Tool GUI on the host desktop (\Cref{fig:power_monitor}).
The MCUs operate at 5V, which is within the USB channel's voltage range of 2.1 V to 5.4 V.
The MCUs consume between 16 and 22 mA of current, for a margin of error of $\sim$2\%.
We report energy consumption for the Arduino Nano 33 BLE (ARMv7E-M Cortex-M4), as it is representative of the other devices.
MM energy measurements are averaged over hundreds of trials.

\begin{figure*}
    \includegraphics[width=0.49\linewidth]{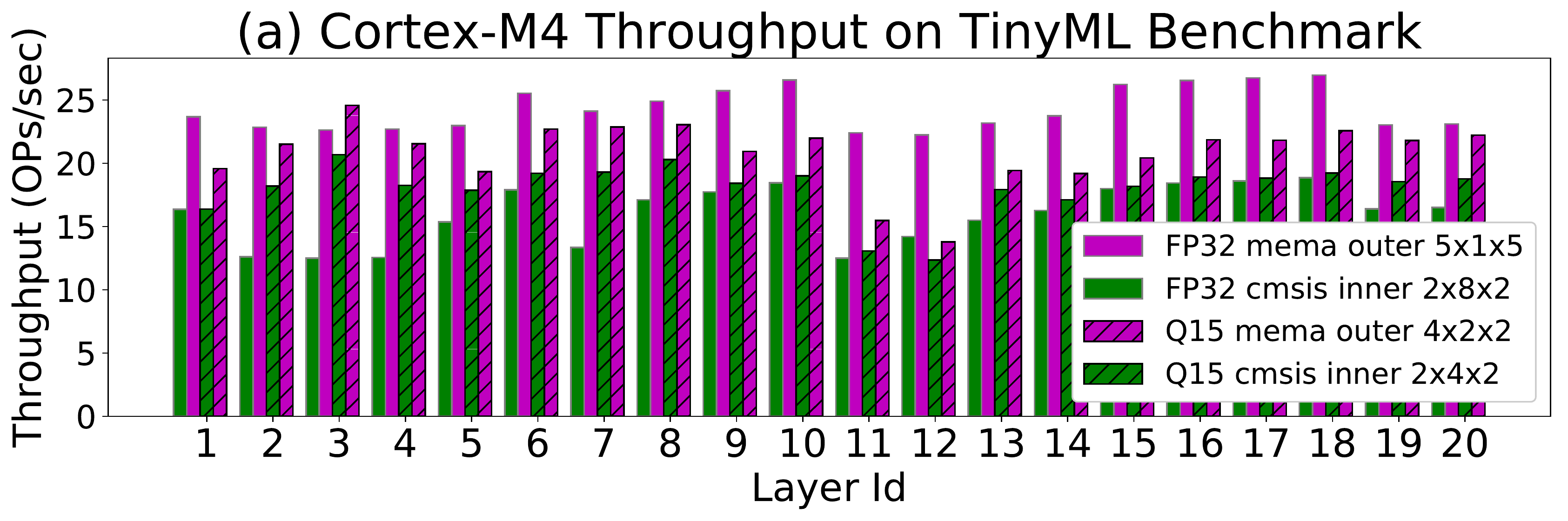}
    \includegraphics[width=0.49\linewidth]{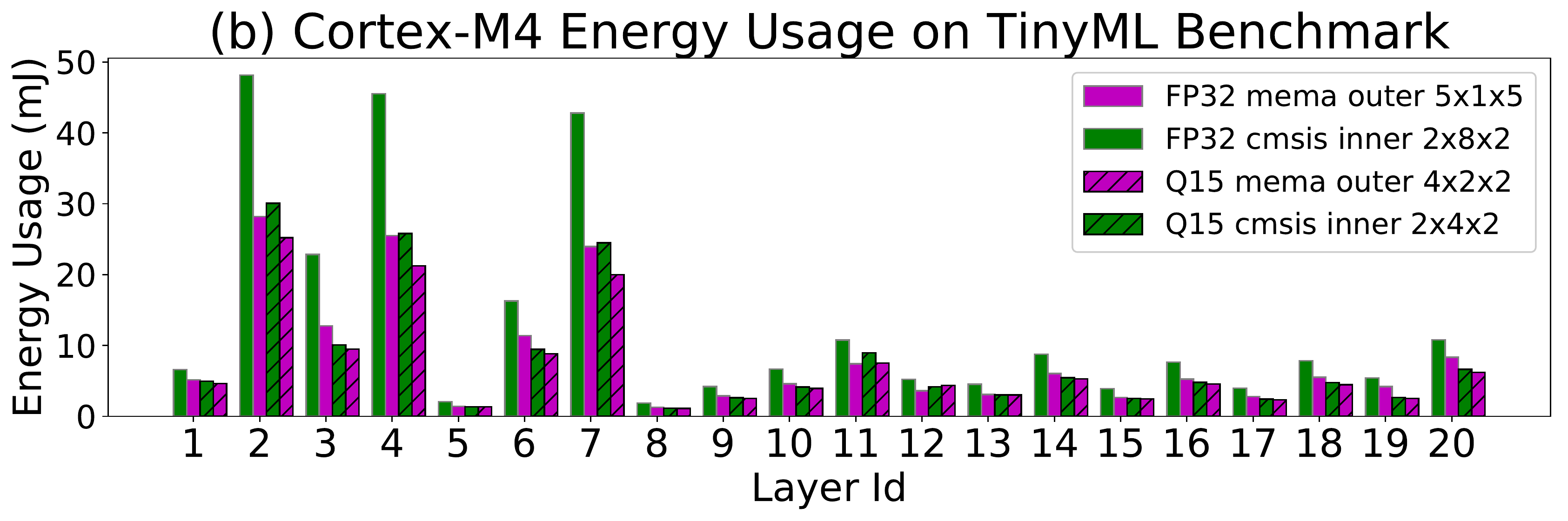} \\
     \begin{minipage}{.44\textwidth}
    \includegraphics[width=\linewidth]{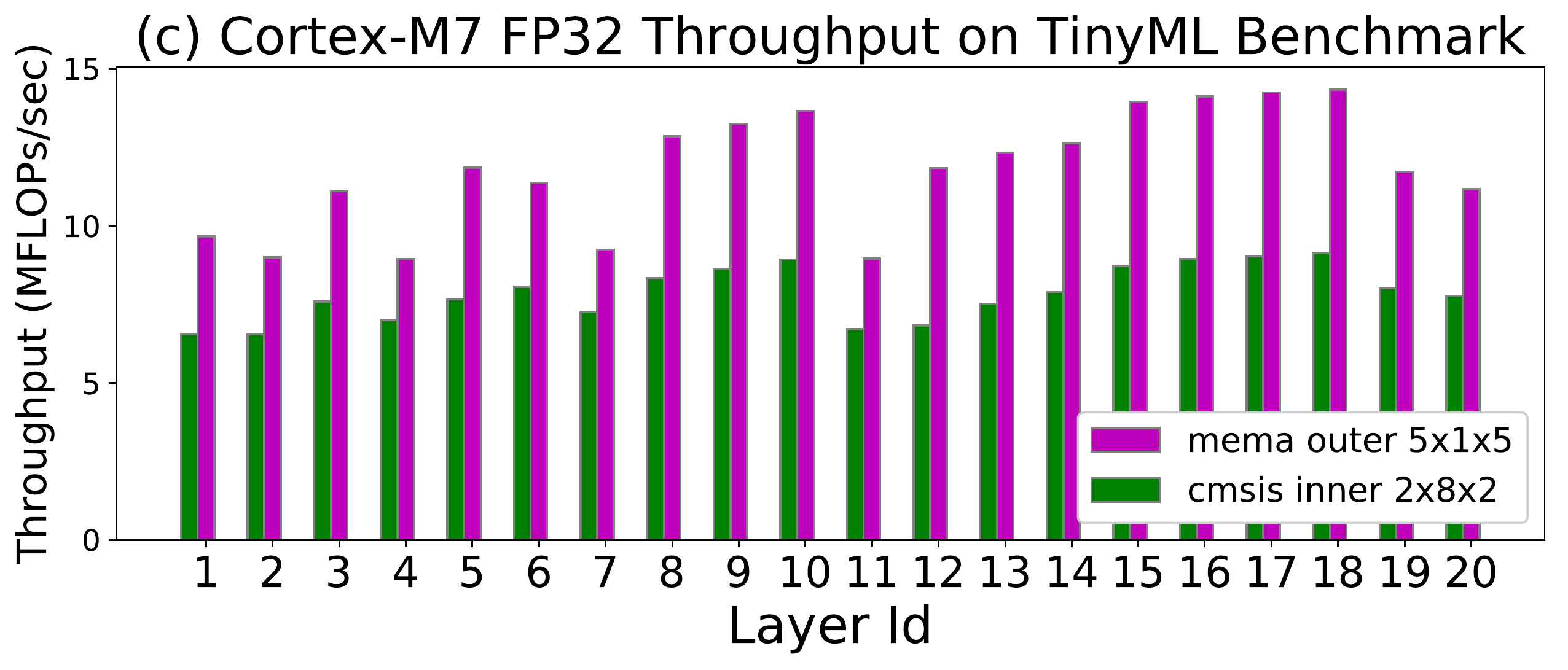}
    \end{minipage}
     \begin{minipage}{.5\textwidth}
     \begin{table}[H]
    \scalebox{0.66}{
   \begin{tabular}{ c l | c l }
\hline
Layer Id & Dimension & Layer Id & Dimension \\
\hline 
1 & $16\times27\times1024$ & 11 & $8\times27\times2304$\\
2 & $16\times144\times1024$ &12 & $16\times8\times2304$\\
3 & $32\times144\times256$ &13 & $32\times16\times576$\\
4 & $32\times288\times256$ & 14 & $32\times32\times576$\\
5 & $32\times16\times256$ & 15 & $64\times32\times144$\\
6 & $64\times288\times64$ & 16 & $64\times64\times144$\\
7 & $64\times576\times64$ & 17 & $128\times64\times36$\\
8 & $64\times32\times64$ & 18 & $128\times128\times36$\\
9 & $64\times40\times122$ & 19 & $256\times128\times9$\\
10 & $64\times64\times125$ & 20 & $256\times256\times9$\\
\hline 
\\
\end{tabular}
}
\label{tab:tinyML}
\end{table}
 \end{minipage}
    \caption{Throughput and energy usage for MEMA and ARM CMSIS-NN on the ARM Cortex-M4 and M7 performing MM with matrices from the TinyML benchmark \cite{banbury2021mlperf}. 
    Each bar group represents the MM for a layer.
    Input matrix dimensions for each layer are shown in the bottom-right table.
    For FP32 MM, MEMA attains up to 1.8x the throughput and 44\% less energy than CMSIS-NN.
    For Q15 arithmetic on the MCU DSPs, only 12 registers are available, limiting tiling options. 
    For Q15 MM, MEMA achieves up to 1.2x the throughput and 20\% less energy than CMSIS-DSP.
    }
    \label{fig:mematinyml}
\end{figure*}

\begin{figure*}
    \centering
    \subfloat{
    \includegraphics[width=0.37\linewidth]{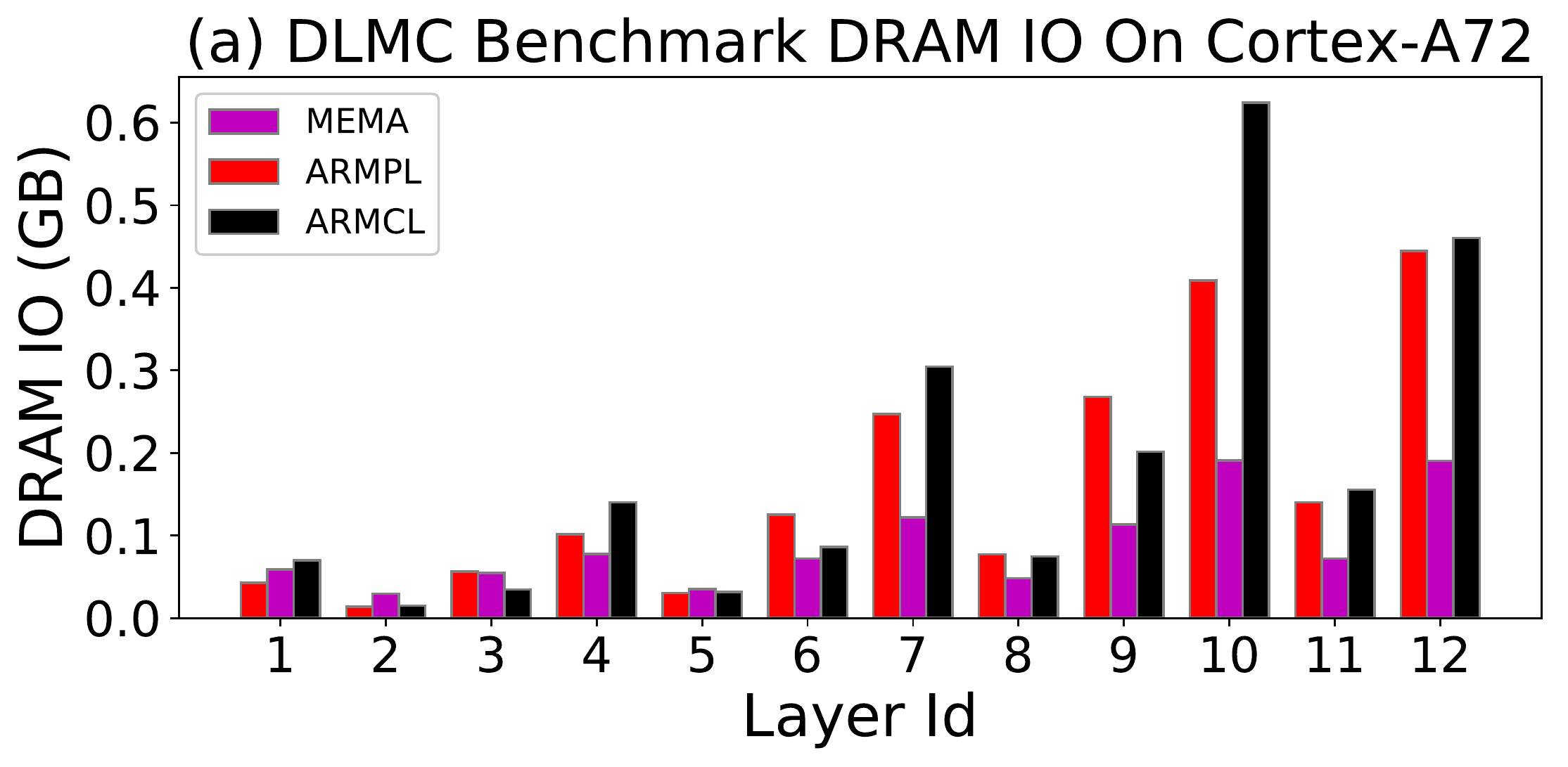}
    }
    \subfloat{
    \includegraphics[width=0.37\linewidth]{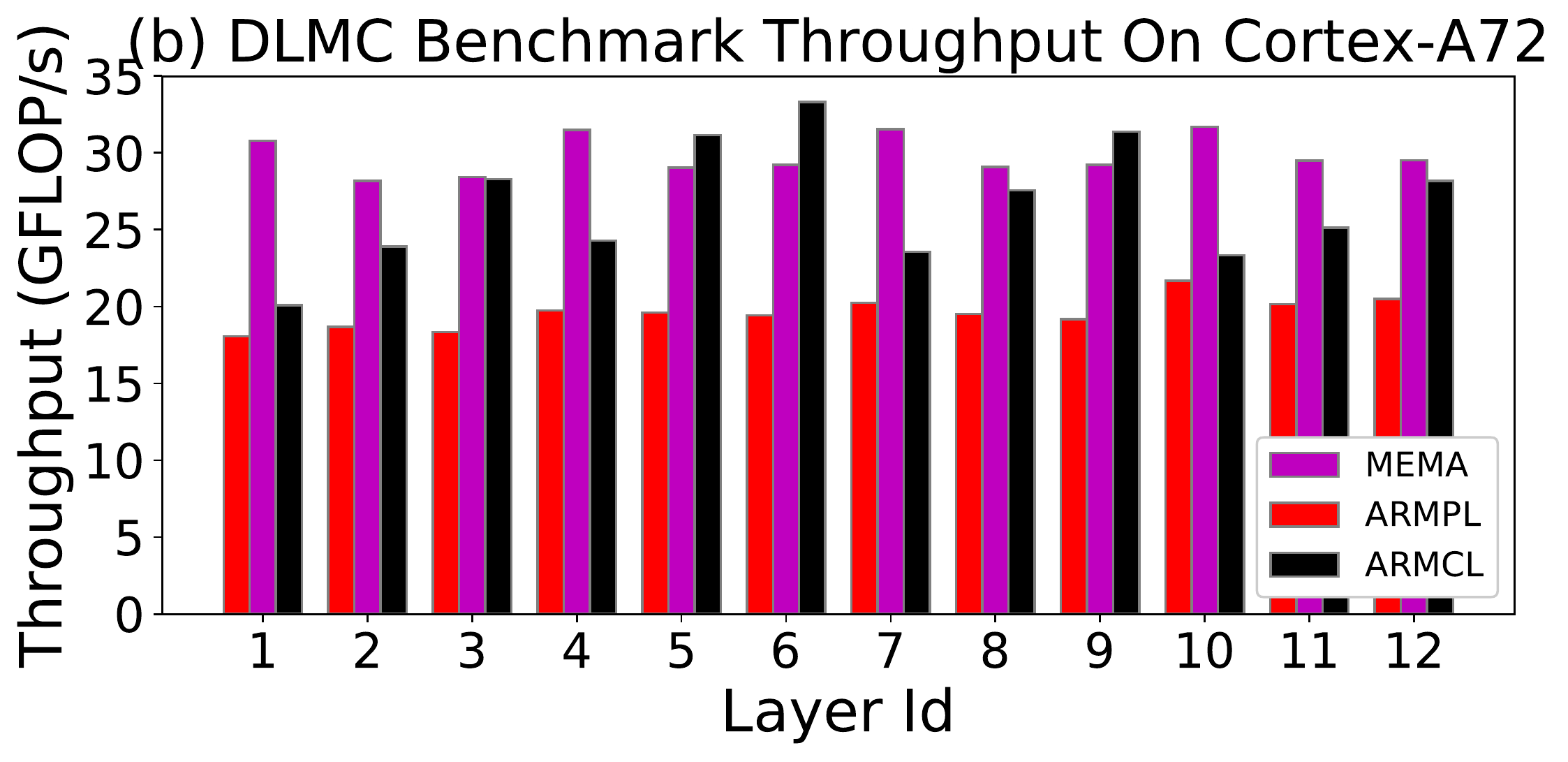}
    }
    \subfloat{
    \scalebox{0.7}{    
    \adjustbox{valign=b}{
    \begin{tabular}{ c l }
\hline
Layer Id & Dimension ($M \times K \times N$) \\
\hline 
1 & $512\times2048\times256$ \\
2 & $512\times512\times256$ \\
3 & $2048\times512\times256$ \\
4 & $512\times2048\times512$ \\
5 & $512\times512\times512$ \\
6 & $2048\times512\times512$ \\
7 & $512\times2048\times1024$ \\
8 & $512\times512\times1024$ \\
9 & $2048\times512\times1024$ \\
10 & $512\times2048\times2048$ \\
11 & $512\times512\times2048$ \\
12 & $2048\times512\times2048$ \\
\hline 
\end{tabular}
}
}
}

    \caption{
    Throughput and external memory IO for MEMA, ARMPL, and ARMCL on the ARM Cortex-A72 when multiplying transformer model matrices from the DLMC Benchmark \cite{dlmc}.
    Input matrix dimensions for each layer are shown in the right-hand table.
    By minimizing external memory accesses (a), MEMA generally achieves peak computation throughput (b). 
    }
    \label{fig:mema_a72}
\end{figure*}

\section{Results and Evaluation}
\label{sec:results}

\subsection{DLMC Benchmark on ARM Cortex-A72}
\label{sec:dlmc_results}

We perform MM on transformer model matrices from the DLMC Benchmark using MEMA, ARMCL, and ARMPL on the ARM Cortex-A72 CPU and collect performance data using \textit{perf} \cite{perf}. 
DRAM accesses are monitored via the ARM PMU event counter (L2 cache refills from DRAM).
The Cortex-A72 contains four cores ($p=4$), a shared L2 cache, and private L1 caches on each core. 
We tile the MM into CB blocks as shown in \Cref{subsec:cake}, where each core computes a $m\times k \times n$ sub-block at a time with $m=k=n$ and $pmk + kpn + p^2mn \leq L2$. 
We further partition the $m\times k \times n$ sub-block into individual $8\times k \times 12$ outer products that can be computed from the 128 32-bit registers available on each core.
The $8\times k \times 12$ shape was chosen by adapting the MCU analysis of \Cref{subsec:tile_selection} to a CPU single core.

ARMCL attains peak throughput on some MM problems (\Cref{fig:mema_a72}b), but may not schedule efficiently.
ARMPL and ARMCL tile MM computations according to the classical Goto's algorithm \cite{gotomain}, which requires more DRAM bandwidth.
In contrast, MEMA attains high throughput for all input shapes by minimizing IO (\Cref{fig:mema_a72}a) via runtime scheduling and CAKE tiling in local memory.

\subsection{MLPerf Tiny Benchmark on ARM Cortex-M4 and M7}
We compare MEMA to ARM CMSIS-DSP on the Cortex-M4 and M7 when performing MM between matrices from the MLPerf Tiny benchmark \cite{banbury2021mlperf}.
CMSIS uses inner product kernels for tiling MM computations such as GEMM-based convolutions. 
For instance, FP32 MM computations are tiled as inner products between $2\times8$ row tiles of $A$ and $8\times2$ column tiles of $B$ to yield $2\times2$ tiles of $C$ (\texttt{cmsis inner 2x8x2} in \Cref{fig:mematinyml}).
In contrast, MEMA tiles MM with outer products between $5\times1$ column vectors of $A$ and $1\times5$ row vectors of $B$ to produce $5\times5$ tiles of $C$ held in registers (\texttt{mema outer 5x1x5} in \Cref{fig:mematinyml}).
MEMA outperforms the ARM kernel by up to $1.8\times$ while reducing energy usage by up to 44\%. 

For saturating fixed-point 16-bit arithmetic (Q15) on the Cortex-M4/7 DSPs, a small local memory (twelve 32-bit registers) limits kernel choices to a few shapes.
Moreover, pairs of values in the input matrices are stored as 32-bit words, requiring pairs of 16-bit values to be unpacked before dual-MAC SIMD computations occur. 
This bit packing stalls the computation by several cycles, resulting in a roofline ridge point that is far to the left.
Despite the severely compute-bound nature of Q15 computations, MEMA improves throughput over CMSIS by reducing external IO.

CMSIS Q15 MM computations are tiled as inner products between $2\times4$ tiles of $A$ and $4\times2$ tiles of $B$ to produce a $2\times2$ tile of $C$ (\texttt{cmsis inner 2x4x2} in \Cref{fig:mematinyml}).
MEMA tiles the MM between $4\times2$ tiles of $A$ and $2\times2$ tiles of $B$ to produce $4\times2$ stationary tiles of $C$ (\texttt{mema outer 4x2x2} in \Cref{fig:mematinyml}).
MEMA and CMSIS use the same SIMD instructions to execute multiple MAC operations per cycle, but MEMA's tiling schedule reduces the number of times $B$ must be streamed from external memory ($\frac{M}{4}$ times for MEMA vs $\frac{M}{2}$ times for CMSIS), shown by the analysis in \Cref{sec:loop_tile_IO}).
Consequently, MEMA attains 20\% higher computation throughput than CMSIS while using up to 23\% less energy.




\section{Conclusion}
We propose the MEMA framework to generate inference runtimes that minimize external memory accesses for TinyML on microcontrollers (MCUs).
By decreasing external memory accesses, MEMA-generated schedules can increase computation throughput and reduce energy consumption.

MEMA can adapt to various TinyML MCUs with different roofline ridge points by selecting data streaming, tile shaping, and loop ordering schemes suited to the hardware (\Cref{fig:tile_mm_and_streaming}b and c).
For example, for neural network benchmarks on the Cortex-M4, the MEMA-generated runtime schedule achieves up to a 1.8x speedup and 44\% less energy than state-of-the-art CMSIS-NN library (\Cref{fig:mematinyml}).
This work demonstrates that simple runtime frameworks, such as MEMA, for TinyML MCUs can significantly reduce memory accesses and energy consumption.

\section{Acknowledgements}
This work was supported in part by the Air Force Research Laboratory under award numbers
FA8750-18-1-0112 and FA8750-22-1-0500, and Meta Platforms Technologies under award
number A51540.


\bibliographystyle{ACM-Reference-Format}
\bibliography{references,refs2}

\end{document}